%% file: main.tex
\pdfoutput=1

\documentclass[11pt]{article}

\usepackage[final]{acl}

\usepackage{times}
\usepackage{latexsym}
\usepackage{booktabs}
\usepackage{arydshln}
\usepackage{listings}
\usepackage{xcolor}  
\usepackage{geometry} 
\definecolor{keywordcolor}{rgb}{0.5, 0.0, 0.5}  
\definecolor{commentcolor}{rgb}{0.0, 0.5, 0.0}  
\definecolor{stringcolor}{rgb}{0.0, 0.0, 0.5}  
\definecolor{backgroundcolor}{rgb}{0.95, 0.95, 0.95}  

\lstdefinestyle{mystyle}{  
    backgroundcolor=\color{backgroundcolor},  
    commentstyle=\color{commentcolor},  
    keywordstyle=\color{keywordcolor},  
    stringstyle=\color{stringcolor},  
    basicstyle=\ttfamily\small \setlength\parindent{0pt},  
    numbers=left,  
    numberstyle=\tiny\color{gray},  
    stepnumber=1,  
    numbersep=7pt,  
    tabsize=2,  
    frame=single,  
    rulecolor=\color{black},  
    breaklines=true,         
    breakatwhitespace=true,  
}  
  
\usepackage[T1]{fontenc}

\usepackage[utf8]{inputenc}
\usepackage{hyperref}

\usepackage{microtype}
\usepackage{enumitem}

\usepackage{inconsolata}

\usepackage{graphicx}
\usepackage{scalerel}
\usepackage{xparse}
\usepackage{multirow}
\usepackage{pifont}
\title{\bench{}: Benchmarking Large Language Models on Dependency Inference with Testable Repositories at Scale}

\author{
    \textbf{Linghao Zhang\footnotemark[1]\textsuperscript{2}},
    \textbf{Junhao Wang\thanks{Work done during internship at Microsoft}\textsuperscript{3}},
    \textbf{Shilin He\thanks{Correspondence to: shilhe@microsoft.com.}\textsuperscript{}},
    \textbf{Chaoyun Zhang\textsuperscript{}},
    \textbf{Yu Kang\textsuperscript{}},
    \textbf{Bowen Li\textsuperscript{4}},
    \\
    \textbf{Jiaheng Wen\footnotemark[1]\textsuperscript{5}},
    \textbf{Chengxing Xie\textsuperscript{4}},
    \textbf{Maoquan Wang\textsuperscript{}},
    \textbf{Yufan Huang\textsuperscript{}},
    \textbf{Elsie Nallipogu\textsuperscript{}},
    \\
    \textbf{Qingwei Lin\textsuperscript{}},
    \textbf{Yingnong Dang\textsuperscript{}},
    \textbf{Saravan Rajmohan\textsuperscript{}},
    \textbf{Dongmei Zhang\textsuperscript{}},
    \textbf{Qi Zhang\textsuperscript{}}
    \\
    \\
    \textsuperscript{}Microsoft,
    \textsuperscript{2}Wuhan University,
    \textsuperscript{3}Tongji University,  \\
    \textsuperscript{4}Shanghai AI Laboratory,
    \textsuperscript{5}Zhejiang University
}

\begin{document}

\input{define}

\maketitle

\input{sections/0_abs}
\input{sections/1_intro}

\input{sections/2_related}

\input{sections/3_dependency_inference}

\input{sections/4_bbb}

\input{sections/5_experiment}
\input{sections/6_results}

\newpage
\bibliography{custom}

\newpage
\appendix
\input{sections/7_appendix}



\end{document}

%% file: define.tex
\newcommand{\he}[1]{\textcolor{red}{HE: #1}}
\newcommand{\jh}[1]{\textcolor{purple}{JH: #1}}
\newcommand{\lh}[1]{\textcolor{blue}{Linghao: #1}}
\newcommand{\FIX}[1]{\textcolor{red}{FIX ME: #1}}
\newcommand{\bench}{\textsc{DI-Bench}}
\newcommand{\baselineall}{All-In-One}
\newcommand{\baselinefile}{File-Iterate}
\newcommand{\baselineimport}{Imports-Only}

\ifdefined\Comment
        \renewcommand{\Comment}[1]{}
\else
        \newcommand{\Comment}[1]{}
\fi

\newcommand{\para}[1]{\vspace{.03in} \noindent\textbf{#1}\hspace{.03in}}

%% file: sections/0_abs.tex
\begin{abstract}

Large Language Models have advanced automated software development, however, it remains a challenge to correctly infer dependencies, namely, identifying the internal components and external packages required for a repository to successfully run. Existing studies highlight that dependency-related issues cause over 40\% of observed runtime errors on the generated repository. To address this, we introduce \bench{}\footnote{Code and data: \url{https://github.com/Microsoft/DI-Bench}}, a large-scale benchmark and evaluation framework specifically designed to assess LLMs' capability on dependency inference. The benchmark features 581 repositories with testing environments across Python, C\#, Rust, and JavaScript. Extensive experiments with textual and execution-based metrics reveal that the current best-performing model achieves only a 42.9\% execution pass rate, indicating significant room for improvement. \bench{} establishes a new viewpoint for evaluating LLM performance on repositories, paving the way for more robust end-to-end software synthesis.

\end{abstract}

%% file: sections/1_intro.tex
\section{Introduction}

Large Language Models (LLMs) have revolutionized automated software development, scaling from function-level code completion~\cite{copilot} to  repository-level code synthesis~\cite{openhands, chatdev, ibrahimzada2024repository}.
\Comment{Recent advancements have enabled these models not only to generate source code but also to orchestrate high-level tasks such as repository configuration, toolchain selection, and code translation.}A pivotal yet often overlooked step in this process is ensuring that generated repositories are fully executable. This requires accurate inference and integration of all necessary dependencies, both internal (across project components) and external (from package ecosystems). Without robust dependency inference, even the most advanced code generation solutions risk failing at runtime, impeding further iteration, evaluation, and reliable deployment. 

\begin{figure}
    \centering
    \includegraphics[width=0.95\linewidth]{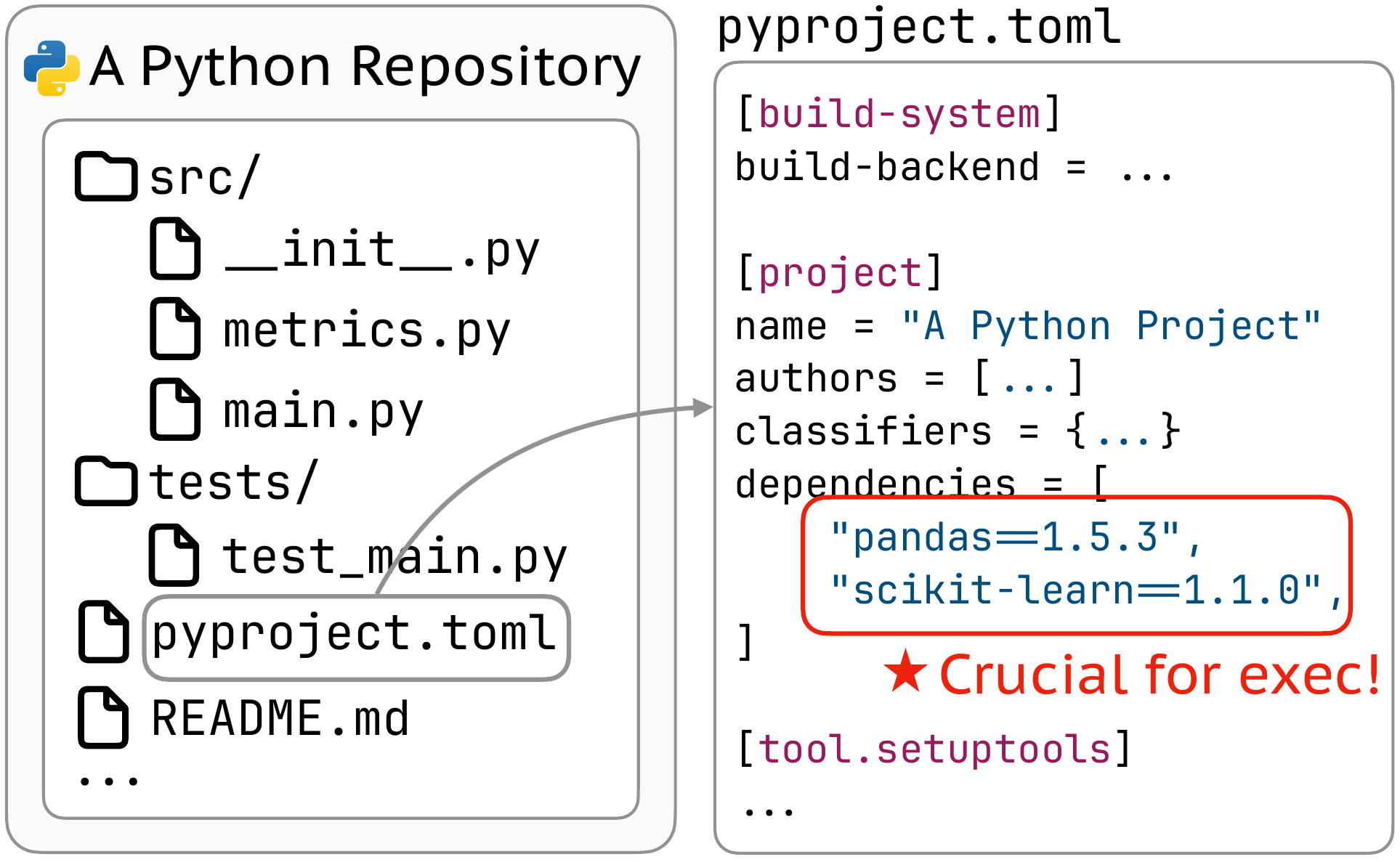}
    \caption{An example of Python project dependencies.}
    \vspace{-1em}
    \label{fig:dep}
\end{figure}

As illustrated in Figure~\ref{fig:dep}, dependency inference involves understanding the intricate relationships within the codebase and mapping out the external packages required for execution. Such dependencies are typically documented in configuration files that may vary from language to language (see Appendix~\ref{appdix:config}). Correctly reconstructing these relationships is a foundational capability: it not only ensures that code generation tools produce functional and self-contained repositories, but it also informs deeper reasoning about project architecture and build systems~\cite{pypi, crates}. Consequently, mastering dependency inference is a critical leap forward for enabling robust, end-to-end software synthesis and maintenance.

Despite the significance of dependency inference, current LLM-based approaches struggle in this area. Works like ChatDev~\cite{chatdev} and DevBench~\cite{devbench}—pioneers in repository-level generation using multi-agent LLM systems—have reported that dependency-related issues (e.g., missing or incorrectly specified modules) account for over 50\% of their observed runtime errors. MetaGPT~\cite{metagpt} also demonstrates that missing or incorrectly generated dependencies represent one of the most significant hallucinations when LLMs attempt to generate the entire project. These challenges highlight the difficulty that state-of-the-art models face in accurately navigating build systems and package repositories. Although existing repository-level benchmarks such as SWEBench~\cite{swebench}, RepoBench~\cite{repobench}, and DevBench~\cite{devbench} offer valuable insights into a model’s ability to handle large contexts and generate code at scale, none focuses on systematically evaluating dependency inference capabilities.

To address this critical gap, we introduce \bench{}, the first comprehensive repository-level benchmark dedicated to dependency inference. \bench{} comprises 581 verified repositories, including 387 regular-sized and 194 large-sized, across four popular programming languages (Python, C\#, Rust, and JavaScript). Each repository is carefully curated to assess a model’s ability to identify both internal and external dependencies. We pair this dataset with a rigorous, multi-faceted evaluation framework. Beyond measuring textual matching accuracy between model-generated and ground-truth dependencies, we propose a novel CI-based execution evaluation by reusing each repository’s intrinsic Continuous Integration (CI) pipelines as automated test harnesses. This approach enables scalable and objective assessment of end-to-end executability, eliminating the costly and error-prone need for manual environment setup.

Through comprehensive experiments involving various LLMs and prompting strategies, we observed that even the best-performing LLM achieved only a 42.9\% executability rate. This finding highlights \textit{significant room for future improvement} in this area. Our analysis revealed that several factors influence performance, including the dependency amount and repository size. Notably, issues such as hallucination and challenges related to dependency metadata emerged as critical bottlenecks that adversely affect model performance.

In summary, our contributions are as follows:
\begin{itemize}[leftmargin=*] 
    \item \textbf{\bench{} Benchmark:} We introduce a pioneering, large-scale, dependency-focused benchmark featuring 581 repositories spanning 4 popular programming languages. It establishes a new standard for evaluating LLMs' capabilities in realistic, repository-scale scenarios. 
    \item \textbf{Dual-Use CI Infrastructure:} We leverage CI workflows not only to identify executable repositories during dataset curation but also to serve as a reliable, fully automated test environment. By using CI pipelines, we ensure that dependency checks remain robust, scalable, and faithful to real-world development practices. 
    \item \textbf{Granular Evaluation Metrics:} We combine coarse-grained runtime executability measures with fine-grained precision and recall on inferred dependencies. This dual-layered approach enables systematic analysis of both functional correctness and textual accuracy with richer insights.
\end{itemize}
By spotlighting dependency inference and offering a dedicated benchmark, our work lays the foundation for advancing LLMs toward robust, end-to-end repository-level software synthesis.

%% file: sections/2_related.tex
\begin{figure*}[t]
    \centering
    \includegraphics[width=0.9\linewidth]{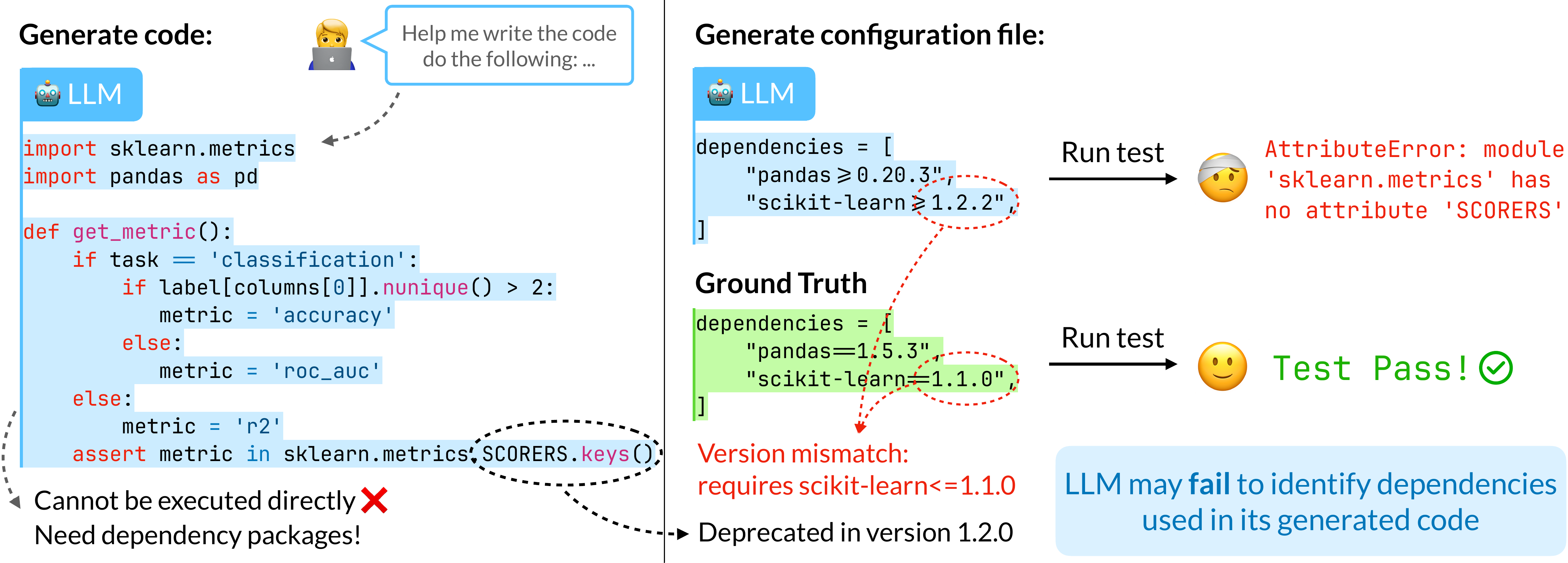}
    \caption{An example of incorrectly identifying dependencies used in code.}
    \label{fig:motivation}
\end{figure*}

\section{Related Works}
Repository-level coding tasks have attracted increasing attention in recent years. Many benchmarks~\cite{repocoder, repobench, cceval} center on code completion tasks at various granularities - from individual lines and API calls to entire function implementations. SWE-Bench and its variant~\cite{swebench, swebench-multimodal} challenge LLMs and LLM-powered systems with real-world scenarios, using issues and pull requests from popular Python repositories on GitHub. While these benchmarks focus on assistant-like tasks, recent work explores LLMs' capabilities in complete project generation. DevBench~\cite{devbench} decomposes the development process into distinct stages and evaluates AI performance at each stage. Agent-As-a-Judge~\cite{agentasjudge} introduces DevAI, innovatively employing LLM agents as evaluators of development outcomes.

However, existing works have not adequately addressed build configuration evaluation: code completion tasks~\cite{repocoder, repobench, cceval} do not include build files in their generation targets, and issue-fixing benchmarks like SWE-Bench~\cite{swebench} contain only 1\% of patches related to build configurations. In repository-level code generation tasks~\cite{devbench, agentasjudge}, build file generation is merely treated as one subtask without specialized evaluation. Prior research in dependency inference~\cite{PyEgo, pigar} has predominantly focused on Python ecosystems using traditional program analysis techniques, while lacking broader language coverage. Our study fills in this gap by providing a benchmark specifically designed for evaluating dependency inference capability across multiple mainstream languages.

%% file: sections/3_dependency_inference.tex
\section{Dependency Inference}
Although many studies focus on repository code generation with LLMs recently, there exists a significant gap between the generated code and the \textit{executable} and \textit{operational} software, \textit{Dependency}. 
%
%
In this paper, we adapt \textit{Dependency Inference}, which aims to generate a list of dependencies based on the source code.  As shown in Figure~\ref{fig:motivation}, the code generated by LLM cannot be executed directly without installing the required dependencies; However, it is non-trivial to identify the correct dependencies using LLMs. The example shows that the LLM generates dependencies with a wrong version (`\textit{scikit-learn==1.1.0}' rather than `\textit{scikit-learn>=1.2.2}'), resulting in execution failure. 

Automatic and accurate dependency inference makes end-to-end code development possible by installing the inferred dependencies for execution. Furthermore, it can enable key scenarios like fully-automated evaluation and iterative code improvement with execution feedback. Besides the repository, dependency inference can be also applied to small code snippets like Python Notebook, incremental code changes, and \textit{etc}.

Formally, the task is formulated as below: Given a software repository containing many source code files and build configuration files where dependency-related sections are masked, the dependency inference task aims to generate a list of inferred dependencies to fill into the configuration.
Formally, we define the task as:
\begin{equation}
    \mathcal{F}: (R, \{b_1^m, b_2^m, ..., b_k^m\}) \rightarrow \{b_1, b_2, ..., b_k\}
\end{equation}
where $R$ denotes the repository including all source files, $b_i^m$ is a build configuration file with dependency masked/removed, $b_i$ is a build configuration with the inferred dependencies. For example, in a Python project, given a \texttt{pyproject.toml} file with masked dependency sections and all source code files, the task is to edit \texttt{pyproject.toml} file to specifying all dependencies required by the project.






%% file: sections/4_bbb.tex
\section{\bench{}}
Focused on the task of \textit{dependency inference}, we introduce \bench{}, a meticulously curated, large-scale benchmark dataset and evaluation framework at the repository level. \bench{} encompasses 581 real-world, testable repository instances across 4 programming languages, providing a comprehensive platform for assessing LLM-based methods in identifying and managing repository dependencies.



\begin{table*}[!ht]
\centering
\caption{Comparison of features between existing benchmarks and \textsc{\bench{}}}
\label{tab:comparison}
\resizebox{2\columnwidth}{!}{
\begin{tabular}{lllllll}
\toprule
\textbf{Benchmark} &\textbf{Task} & \textbf{Evaluation} & \textbf{Scope} & \textbf{Languages} & \textbf{\#Repo} & \textbf{Curation} \\ 
\midrule
MBPP~\cite{mbpp} & Code Generation & Unit Tests & Function & Python & N/A & Manual \\
HumanEval~\cite{humaneval} & Code Generation  & Unit Tests & Function & Python & N/A & Manual\\
ClassEval~\cite{classeval} & Code Generation & Unit Tests & Class & Python & N/A & Manual \\
RepoEval~\cite{repocoder} & Code Completion & Textual \& Unit Tests & Repo-level & Python & 14 & Manual \\
RepoBench~\cite{repobench} & Retrieval \& Completion & Only Textual & Repo-level & Python, Java & 1,669 & Automated \\
CrossCodeEval~\cite{cceval} &Code Completion & Only Textual & Repo-level & Python, Java, C\#, TS & 1,002 & Automated \\
EvoCodeBench~\cite{evocodebench} & Code Generation & Unit Tests & Repo-level & Python & 25 & Automated \\
RepoMasterEval~\cite{repomastereval} & Code Completion & Unit Tests & Repo-level & Python, TS & 6 & Manual \\ 

\midrule

\bench{} & \textbf{Dependency Inference} & \textbf{Textual \& Test Suite} &\textbf{Repo-level} & \textbf{Python, Rust, C\#, JS} & \textbf{581} & \textbf{Automated}\\

\bottomrule
\end{tabular}
}
\end{table*}

\subsection{Statistics \& Features}

\bench{}'s instances, sourced from real-world repositories, are categorized into two subsets based on repository size: \textit{regular} and \textit{large}. The \textit{regular} subset includes repositories with fewer than 120k tokens\footnote{Token counts are calculated with Llama 3.2 tokenizer.}, ensuring they fit within the context length limits of recent LLMs. It comprises 387 instances with an average of 11.9 dependencies. The \textit{large} subset consists of 194 repositories with more than 120k tokens and the average dependency count is 29.7. Table~\ref{tab:stats} provides detailed statistics of \bench{}, while Figure~\ref{fig:distri} illustrates the overall distribution of token and dependency counts using Kernel Density Estimation (KDE) curves. The dataset exhibits a wide size distribution, with smaller repositories being more prevalent.
Table~\ref{tab:comparison} shows a comparative analysis of features distinguishing \bench{} from existing code task benchmarks. The unique attributes of \bench{} include:


\begin{table}[t]
\centering
\caption{Statistical summary of \bench{}}
\label{tab:stats}
\resizebox{\columnwidth}{!}{
\begin{tabular}{llrrrrr}
\toprule
Subset & Lang & \#Files & \#LoC & \#Tokens & \#Deps. & \#Tests\\
\midrule
\multirow{5}{*}{Regular} & Python & 30.9 & 3.0K & 31K & 5.5 & 47.1 \\ 
    & Rust  & 20.2 & 3.4K & 32K & 10.8 & 21.0 \\
    & C\# & 69.7 & 4.1K & 39K & 25.9 & 30.6 \\
    & JS & 14.9 & 1.6K & 15K & 5.7 & 42.7 \\
    \cdashline{2-7}   & \textit{Avg.} & \textit{33.9} & \textit{3.0K} & \textit{29K} & \textit{11.9} & \textit{35.3} \\
\midrule
\multirow{5}{*}{Large} & Python & 268.3 & 45.6K & 519K & 11.8 & 547.3 \\ 
    & Rust  & 95.0 & 23.9K & 283K & 45.4 & 155.4 \\
    & C\# & 353.9 & 37.3K & 369K & 44.5 & 132.6 \\
    & JS & 144.9 & 26.0K & 367K & 16.2 & 304.4 \\
    \cdashline{2-7} & \textit{Avg.} & \textit{218.0} & \textit{33.4K} & \textit{385K} & \textit{29.7} & \textit{285.1} \\
\bottomrule
\end{tabular}
}
\end{table}


\para{Beyond Code.} \bench{} focuses on a crucial challenge in real-world software development: dependency inference. This essential aspect is often overlooked in existing studies.

\para{Test Execution.} \bench{} not only evaluates result correctness through textual matching but also executes project test suites, providing a straightforward and reliable evaluation.

\para{Practical and Verified.} The repository instances included in \bench{} are sourced from real-world projects on GitHub, thus making the benchmark both practical and challenging. Each project undergoes verification to ensure its validity.

\para{Diverse Long Inputs.} The dataset includes two subsets, regular and large, with a wide distribution of context lengths, ranging from small repositories with a few files to large projects with over 200 files.

\para{Continually Updatable.} We have developed a dataset curation pipeline that is fully automated, scalable, and continuously updatable, eliminating the need for manual annotation to set up environments and run tests.

\para{Open Solution.} Our evaluation framework features two complementary datasets: while both the regular and large sets welcome various approaches including language models and agentic systems, the large set presents additional challenges of model context limits, specifically motivating the exploration of novel methodologies.

\subsection{Dataset Construction}

\begin{figure*}[ht]
    \centering
    \includegraphics[width=0.9\linewidth]{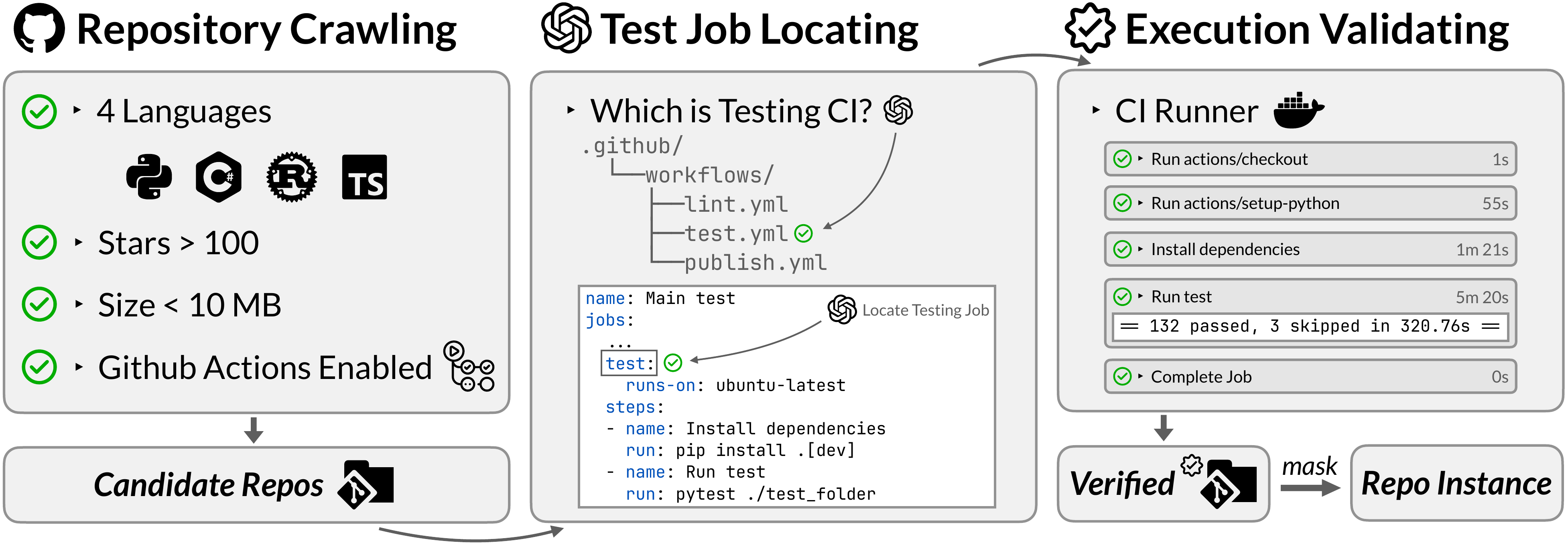}
    \caption{CI-based curation pipeline for \textsc{\bench{}}.}
    \label{fig:curating}
\end{figure*}

Creating a dataset that supports execution-based evaluation at the repository level is challenging. Previous works often involve manual setting up environments and writing test scripts, which can require significant human and engineering effort and cannot scale up to larger datasets. As shown in Table~\ref{tab:comparison}, the existing largest repository-level benchmark supporting test execution contains only 25 repositories. To address this, we leverage GitHub Actions~\cite{actions}—a widely used continuous integration (CI) tool that allows developers to automate test execution through YAML configuration files. By reusing these developer-written CI workflows within repositories, we propose an automated curation pipeline that eliminates human engagement during the benchmark construction, ultimately resulting in a dataset of 581 testable repositories—23 times larger than the largest previous benchmark. With the large-scale dataset, we can provide more generalizable insights and more robust evaluations. Figure~\ref{fig:curating} illustrates steps to construct \bench{} with details listed below.

\paragraph{Repository Crawling.}
The goal of this phase is to collect GitHub repositories that meet the following criteria: 1) Written in one of the four programming languages: Python, C\#, Rust, or JavaScript (the characteristics of these languages and their dependency configurations are detailed in Appendix~\ref{appdix:config}). These languages are popular, possess a standardized dependency packages ecosystem, and have clear standards for specifying dependencies. 2) Have more than 100 stars, serving as a quality filter criterion. 3) Repository size is less than 10MB to avoid extremely large repositories and maintain a manageable dataset size. 4) Most importantly, the repository must have GitHub Actions enabled, indicated by the presence of the \texttt{.github/workflows} folder. Repositories that meet these criteria proceed as candidate repositories into subsequent phases.

\paragraph{Test Job Locating.}
Repositories often define multiple workflows to perform tasks unrelated to testing, such as linting and publishing. These tasks may also be defined within different jobs in the same workflow configuration file. Due to the lack of a specific naming convention, we introduce an LLM-assisted locating process to identify the specific jobs responsible for executing project tests. At the execution stage, only the test job will be run.

\paragraph{Execution Validating.}
We use \texttt{act}~\cite{act} as a local runner for GitHub Actions, enabling local execution of repository testing CI. In this phase, by executing the test jobs of candidate repositories, we obtain those that successfully follow the workflow and pass all tests as expected. The validation phase ensures that the selected repositories are correct and executable, highlighting the advantage of our proposed CI-based testing approach: fully automated and scalable.

\paragraph{Dependency Masking.}
After the validation, we utilized an automated script to remove the sections specifying dependencies in the configuration files. We further performed sanitization by removing any existing dependency lock files (e.g., in JavaScript) to prevent potential ground truth leakage and ensure proper execution. This process ultimately produced the instances included in \bench{}.

%% file: sections/5_experiment.tex
\section{Experiment Setup}
This section provides a detailed description of the experimental settings, including the LLMs, baseline methods, and evaluation metrics.


\paragraph{Baseline Methods.}
We designed three baseline systems with various prompting strategies to evaluate how LLMs perform in the \textit{dependency inference} task, intentionally avoiding complex techniques such as agent-based methods.

\begin{itemize}[leftmargin=*]
    \item \baselineall{}: The approach concatenates all the source code of a repository into a single query for model generation. It serves as a straightforward yet computationally intensive baseline.
    \item \baselinefile{}: The method processes each individual file in the repository to generate dependencies, with the results subsequently aggregated to feed into the model for generating the final output. This simulates a modular and distributed reasoning approach.
    \item \baselineimport{}: The approach collects all import-related statements from the code base as the input context to LLMs using \textit{tree-sitter}~\cite{treesitter}. For Python and JavaScript, we extract all \texttt{import} statements; for C\# and Rust, we extract all \texttt{use} statements. More details about tree-sitter are provided in the Appendix~\ref{appdix:treesitter}. 
\end{itemize}




\begin{table*}[ht]  
\centering  
\caption{Performance of benchmark methods across programming languages and repository sizes on GPT-4o, where Exec denotes the executability rate, P/R/F1 denote Precision, Recall, and F1-score, FR denotes Fake Rate, which is the lower, the better. Note that the Large repositories cannot fit into the \baselineall{} method (denoted with `-').}  
\label{tab:baseline-performance-repo-size}  
\resizebox{0.8\textwidth}{!}{  
\begin{tabular}{llcccccccccc}  
\toprule  
\multirow{2}{*}{Lang} & \multirow{2}{*}{Method} & \multicolumn{5}{c}{Regular} & \multicolumn{5}{c}{Large} \\  
\cmidrule(lr){3-7} \cmidrule(lr){8-12}  
         &              & Exec & P & R & F1 & FR & Exec & P & R & F1 & FR \\  
\midrule  
\multirow{3}{*}{Python} & \baselineall{}  & 42.9 & 61.8 & 73.6 & 67.2 & 2.8 & - & - & - & - & -\\  
         & \baselinefile{}   & 29.6 & 38.1 & 75.6 & 50.7 & 4.5 & 8.0 & 19.5 & 35.3 & 25.1 & 6.4 \\  
         & \baselineimport{} & 36.7 & 56.5 & 74.9 & 64.4 & 3.9& 18.0 & 36.9 & 46.9 & 41.3 & 23.1 \\  
\midrule  
\multirow{3}{*}{Rust}   & \baselineall{}   & 11.2 & 93.7 & 74.4 & 82.9 & 0.8 & - & - & - & - & -\\  
         & \baselinefile{}    & 7.1 & 74.7 & 75.6 & 75.2 & 1.1  & 2.0 & 45.0 & 68.8 & 54.4 & 6.3  \\  
         & \baselineimport{}- & 4.1 & 88.9 & 65.0 & 75.1 & 1.0   & 2.0 & 84.9 & 50.6 & 63.4 & 12.4 \\  
\midrule    
\multirow{3}{*}{C\#}    & \baselineall{}   & 13.5 & 59.9 & 39.4 & 47.5 & 3.7 & - & - & - & - & - \\  
         & \baselinefile{}    & 5.2 & 27.7 & 34.1 & 30.6 & 6.8  & 0.0 & 20.4 & 33.0 & 25.2 & 6.0  \\  
         & \baselineimport{}  & 3.1 & 52.7 & 29.8 & 38.1 & 5.2 & 0.0 & 49.1 & 19.2 & 27.6 & 6.3 \\  
\midrule  
\multirow{3}{*}{JavaScript}  
         & \baselineall{}  & 43.2 & 86.9 & 67.7 & 76.1 & 4.8 & - & - & - & - & - \\  
         & \baselinefile{}    & 32.6 & 52.2 & 62.6 & 57.0 & 2.9 & 15.6 & 34.2 & 52.1 & 41.3 & 2.7 \\   
         & \baselineimport{} & 22.1 & 73.6 & 46.7 & 57.1 & 6.2 & 6.7 & 55.4 & 15.5 & 24.2 & 2.0 \\  
\bottomrule
\end{tabular}}  
\vspace{-1em}
\end{table*} 

\paragraph{Metrics.} we use textual and execution-based metrics, and the fake rate in evaluation.

\begin{itemize}[leftmargin=*]
\item Textual Accuracy: This metric assesses whether the generated dependencies align with the ground truth from a textual matching perspective.
We compute the 
\textit{Precision} (ratio of correct dependencies among model-generated ones),  \textit{Recall} (Ratio of correct dependencies among all ground-truth ones), and \textit{F1} (The harmonic mean of the above).
\item Executability Rate: This metric measures whether the project can be successfully built and executed through CI testing pipeline with the generated dependencies. A score of 1 is assigned if all tests passed successfully; otherwise, a score of 0 will be given. Whether the tests pass is the most direct and reliable indicator of the correctness of the generated dependencies.
\item Fake Rate: This metric represents the proportion of the generated dependencies that cannot be found in the package ecosystem (for external dependencies) or in the local repository directory (for internal dependencies). It highlights the hallucination issue in LLMs, where non-existent dependencies or versions are generated.
\end{itemize}

\paragraph{Models.}
Since code repositories are usually very long, we choose popular LLMs that support 128k context windows as the backbone models, including GPT-4o~\cite{gpt4o}, GPT-4o-mini~\cite{gpt4omini}, Llama 3.1-Instruct~\cite{llama31}, DeepSeek-Coder-V2-Lite-Instruct (MoE)~\cite{deepseek} and Qwen-Coder-V2.5-Instruct series~\cite{qwen2}.

%% file: sections/6_results.tex
\section{Experimental Results}
\label{sec:results}

\subsection{Performance of Baseline Methods}
\label{sec:baseline_results}
We start by conducting preliminary experiments utilizing three baseline systems — \baselineall{}, \baselinefile{}, and \baselineimport{} on GPT-4o and GPT-4o-Mini (Table~\ref{tab:baseline-performance-repo-size-2} in Appendix ~\ref{appendix:baseline-result}), encompassing both the regular and large subsets. Table~\ref{tab:baseline-performance-repo-size} shows the results with several key insights:

\paragraph{Challenging Nature of Dependency Inference} Dependency inference presents a significant challenge for contemporary LLMs. In the regular subset (< 120k tokens), even the best-performing models achieved executability rates of below 50\% for scripting languages such as Python and JavaScript and only around 10\% for compiled languages like Rust and C\#. These findings underscore the limitations of current models in accurately inferring dependencies in various languages.

\begin{table*}[t]
\centering
\caption{Model performance across programming languages with the \baselineall{} approach on \bench{}.}
\label{tab:lang-performance}
\resizebox{0.8\textwidth}{!}{
    \begin{tabular}{llllcccccc}
    \toprule  
    Language & Model & Size & Exec & P & R & F1 & FR \\
    \midrule 
    Python 
     & GPT-4o & - & 42.9 & 61.8 & 73.6 & 67.2 & 2.8 \\
     & GPT-4o-mini & - & 24.5 & 56.5 & 57.5 & 57.0 & 2.0 \\     
     & Llama-3.1-8B-Instruct           & 8B       & 13.3 & 28.8 & 38.4 & 32.9 & 4.3 \\
     & DeepSeek-Coder-V2-Lite-Instruct & 16B(MoE) & 17.3 & 48.0 & 48.6 & 48.3 & 18.4 \\
     & Qwen2.5-Coder-7B-Instruct       & 7B       & 22.4 & 55.4 & 44.7 & 49.5 & 5.3 \\
    \midrule 
    Rust 
     & GPT-4o & - & 11.2 & 93.7 & 74.4 & 82.9 & 0.8 \\
     & GPT-4o-mini & - & 7.1 & 76.0 & 49.1 & 59.7 & 1.0 \\
     & Llama-3.1-8B-Instruct           & 8B       & 1.0 & 58.2 & 37.0 & 45.2 & 10.7 \\
    & DeepSeek-Coder-V2-Lite-Instruct & 16B(MoE) & 2.0 & 75.9 & 40.8 & 53.0 & 2.8 \\
      & Qwen2.5-Coder-7B-Instruct & 7B & 6.1 & 71.3 & 41.0 & 52.0 & 2.0 \\
    \midrule 
    C\# 
     & GPT-4o & - & 13.5 & 59.9 & 39.4 & 47.5 & 3.7 \\
     & GPT-4o-mini & - & 4.2 & 41.5 & 22.3 & 29.0 & 11.9 \\
     & Llama-3.1-8B-Instruct           & 8B       & 0.0 & 15.2 & 7.6 & 10.1 & 23.2 \\
     & DeepSeek-Coder-V2-Lite-Instruct & 16B(MoE) & 1.0 & 33.6 & 7.4 & 12.1 & 9.2 \\
     & Qwen2.5-Coder-7B-Instruct       & 7B       & 1.0 & 22.7 & 17.1 & 19.5 & 14.9 \\
    \midrule JavaScript 
     & GPT-4o & - & 43.2 & 86.9 & 67.7 & 76.1 & 4.8 \\  
     & GPT-4o-mini & - & 16.8 & 84.6 & 31.6 & 46.0 & 2.5 \\
     & Llama-3.1-8B-Instruct           & 8B      & 9.5 & 67.2 & 16.7 & 26.8 & 1.5 \\
     & DeepSeek-Coder-V2-Lite-Instruct & 16B(MoE) & 17.9 & 83.9 & 32.0 & 46.3 & 2.4 \\
     & Qwen2.5-Coder-7B-Instruct       & 7B       & 16.8 & 81.8 & 43.3 & 56.6 & 3.5 \\
    \bottomrule
    \end{tabular}
    }
\end{table*}

\paragraph{Impact of Repository Size}\label{para-impact-repo-size} Large repositories, characterized by extensive contexts and complex dependency structures, are more challenging for dependency inference. Executability rates in the large subset were markedly lower across all baseline methods compared to the regular subset. For \baselinefile{} and \baselineimport{}, the performance gap was especially evident, indicating the difficulty of adapting these methods to large repositories. 

\paragraph{Importance of Models and Prompting Strategies} The choice of backbone LLMs and the construction of prompts play crucial roles in determining performance outcomes. For instance, the \baselineall{} approach with GPT-4o, by merging the entire code base into a single query, consistently outperformed other methods on executability. However, this approach does not work for larger repositories. While the \baselinefile{} and \baselineimport{} methods can process large repositories, their performance significantly declined without the full code context. The finding reveals the trade-off between prompting strategies and repository sizes to achieve optimal performance on the task. 

\paragraph{Hallucination Issues} A recurring issue across all methods was the generation of hallucinated dependencies, i.e., non-existent packages or versions, as indicated by the Fake Rate. Specifically, we observed a remarkably higher Fake Rate on large subset with \baselineimport{} method.  
In Section~\ref{sec:ablation}, we will show that the hallucination adversely affected the executability. 

\subsection{Performance of Varying Models}
Due to space constraints, we focus on reporting benchmarking results on the \bench{} Regular dataset using the \baselineall{} approach in the following sections. This method was selected due to its superior performance as shown in Table~\ref{tab:baseline-performance-repo-size}, as well as its simplicity and capability to reflect a zero-shot setting for dependency inference.

Specifically, in this section, we evaluate various models and report the results in Table~\ref{tab:lang-performance}. The table reveals that open-sourced models achieve significantly lower performance compared to models like GPT-4o. Notably, the Qwen-7B model demonstrates superior performance across all metrics, outperforming the other two open-sourced models and occasionally even surpassing the GPT-4o-mini model. Additionally, we find that although the open-sourced models can achieve moderate textual accuracy, their executability rate is considerably lower. This highlights the challenge of not only understanding and generating correct textual dependencies but also ensuring that the generated code can be executed successfully.

In addition, we vary the model sizes of the QWen2.5-Coder-Instruct series with the \baselineall{} method, where the model size ranges from 3B, 7B, 14B to 32B. We observed that the model in general achieves better performance when increasing the model size. The results and more analysis are presented in Appendix~\ref{appendix:qwen-various-size}


\begin{figure}
    \centering
    \includegraphics[width=\linewidth]{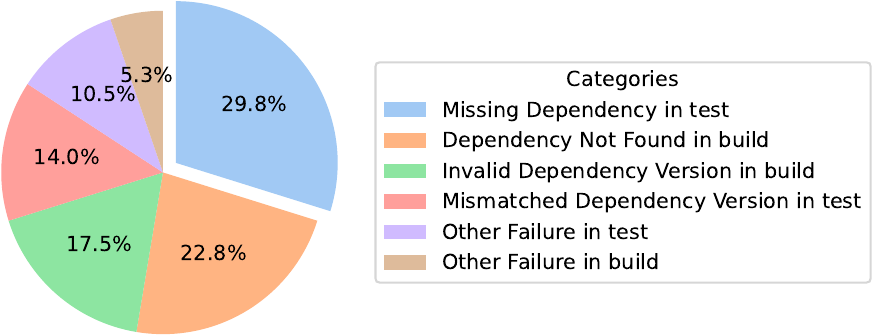}
    \vspace{-2em}
    \caption{Distribution of failure categories (GPT-4o, \baselineall{} setting, Python).}
    \vspace{-1em}
    \label{fig:categories}
\end{figure}

\para{Failure Categories and Distribution.} To better understand why the execution failed with model-generated dependencies, we manually analyzed the failure cases of GPT-4o, the best-performing model, under the \baselineall{} setting in Python. 
As shown in Figure~\ref{fig:categories}, the most common failure category is ``\textit{Missing Dependency in Test}'', which means that the model missed to generate some dependencies that are required during the testing evaluation. Additionally, ``\textit{Dependency Not Found in Build}'' and ``\textit{Invalid Dependency Version in Build}'' also account for a significant proportion. These indicate that the model-generated dependency specifications either include nonexistent packages or specify nonexistent versions, leading to failures when installing generated dependencies.

\subsection{Further Analysis and Ablation Study}
\label{sec:ablation}
In this section, we conduct further analysis with focuses on how repository size and the amount of dependencies affect dependency inference performance, as well as the impact of dependency metadata and the hallucination issue.

\begin{table}[t]
\centering
    \caption{Execution success improvement by replacing predicted dependency metadata with oracle metadata.}
    \label{tab:metadata-impact}
    \resizebox{0.9\linewidth}{!}{%
    \begin{tabular}{llcr}
        \toprule
        \textbf{Language} & \textbf{Exec} & \textbf{Exec (with Orac.)}  & $\Delta $ \\
        \midrule
        Python & 42.9 & 55.1  & +28.4\% \\
        Rust & 11.2 & 38.8 & +246.4\% \\
        C\# & 13.5 & 15.6 & +15.6\% \\
        JavaScript & 43.2 & 67.4 & +50.5\% \\ 
        \bottomrule
    \end{tabular}%
    }
\end{table}

\begin{figure}[!ht]
    \includegraphics[width=1\linewidth]{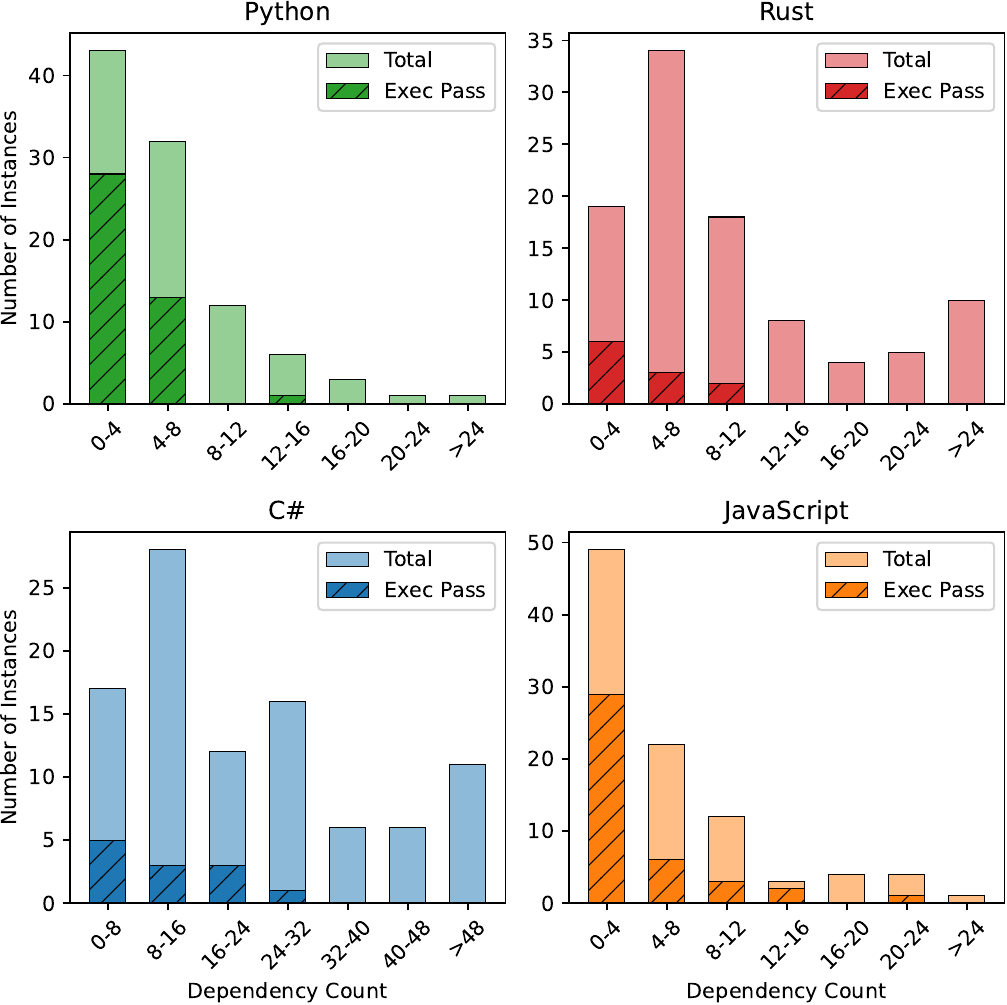}
    \caption{Execution pass rate w.r.t dependency count.}
    \label{fig:exec_dep}
\end{figure}

\paragraph{Challenges in dependency inference for larger repositories with more dependencies.} 
As illustrated in Figure~\ref{fig:exec_dep}, inference accuracy decreases significantly as the number of dependencies grows. This trend is consistent across all languages, particularly those with complex dependency structures like Rust and JavaScript. The decline in performance is attributed to the difficulty of maintaining accurate dependency mappings as their quantity increases, highlighting spaces for future enhancement of LLMs.
Besides, we made further analysis about how the repository size affect the performance on the regular dataset and results are depicted in Figure~\ref{fig:exec_token} (Appendix~\ref{appendix:repo_size}). We observed a negative correlation between repository size and model performance, which aligns with the finding obtained in Table~\ref{tab:baseline-performance-repo-size}. This suggests that long-context reasoning~\cite{ruler, longbench} remains a significant challenge for LLMs, as longer input contexts lead to increased complexity.



\para{Reasoning the dependency metadata is a bottleneck.}
In previous experiments, we found that while textual accuracy was relatively high, the executability rate was significantly lower. For example, GPT-4o achieved a precision of 61.8\% and recall of 73.6\% on Python, while the executability rate was only 42.9\%. We suspect this discrepancy arises from incorrect metadata generation in dependencies, such as package version constraints, extra features and so on, (examples can be found in Appendix~\ref{appdix:config}).  To validate this hypothesis, we replaced the predicted dependencies with oracle metadata and observed a notable increase in the executability rate. As shown in Table~\ref{tab:metadata-impact}, the Python executability rate improved from 42.9\% to 55.1\%, representing a relative increase of 28.4\%. It  demonstrates the importance of accurate dependency metadata for successful execution of dependency configurations.

%

\para{Hallucination hurts the executability. } We observed all models generate hallucinated dependencies that do not exist, as indicated by the fake rate. 
Although the fake rate was relatively low, excluding the hallucinated dependencies can  improve the executability, as shown in Table~\ref{tab:hallucination-impact}. These improvements, though modest, reinforce the need for more accurate dependency predictions. Hallucination issues remain one of the primary obstacles to improving the reliability of dependency inference.

\begin{table}[t]
\centering
    \caption{Impact of hallucination on exeutability rate.}
    \label{tab:hallucination-impact}
    \resizebox{0.9\columnwidth}{!}{%
    \begin{tabular}{lccr}
        \toprule
        \textbf{Language} & \textbf{Exec} & \textbf{Exec w.o. Fake Dep.} & \textbf{$\Delta$} \\
        \midrule
        Python & 41.8 & 43.8 & +4.8\%\\
        Rust & 11.2 & 13.3 & +18.8\%\\
        C\# & 12.5 & 13.5 & +8.9\%\\
        JavaScript & 43.2 & 43.2 & +0\%\\ 
        \bottomrule
    \end{tabular}%
    }
\end{table}


\Comment{
\begin{table}[t]  
\centering  
\caption{Model Performance Across Programming Languages for Qwen2.5-Coder-Instruct}  
\label{tab:size-performance}  
\resizebox{\columnwidth}{!}{  
\begin{tabular}{llccccccc}  
\toprule  
Language & Size & Exec & P & R  & F1 & FR \\  
\midrule  
{Python}
         & 3B   & 18.0          & 34.2          &  51.1          & 41.0  & 6.5 \\
         & 7B   & 22.0          & 53.4          &  42.1          & 47.1  & 5.3          \\
         & 14B   & 30.0 & 51.2 & 63.5 & 56.7 & 5.0 \\
         & 32B  & 35.0          & 54.5          &  67.5          & 60.3  & 5.7  \\
\midrule  
{Rust}  
         & 3B  & 2.0           & 58.2           &  57.4         &  57.8 & 7.0 \\
         & 7B  & 6.0 & 67.3 & 40.2 & 50.3 & 2.1          \\  
         & 14B  & 11.0 & 80.6 & 64.5 & 71.7 & 1.5 \\
         & 32B & 10.0 & 85.2 & 73.3 & 78.8 & 1.7 \\
\midrule  
{C\#}  
         & 3B   & 0.0 & 18.4 & 4.4 & 7.1 & 19.2 \\
         & 7B   & 1.0 & 22.6 & 17.2 & 19.6 & 14.7 \\
         & 14B  & 5.0 & 36.0 & 32.7 & 34.3 & 7.1 \\
         & 32B  & 12.0 & 35.2 & 35.8 & 35.5 & 8.9 \\
\midrule  
{JavaScript}  
         & 3B   & 4.0  & 68.8 & 3.0 & 5.8 & 0.0           \\
         & 7B  & 17.0 & 53.4 & 42.1 &  46.9  & 5.3          \\
         & 14B & 32.0 & 84.5 & 59.3 & 69.7 & 0.3 \\
         & 32B  & 36.0 & 77.1 & 60.2 & 67.6 & 10.5 \\
\bottomrule  
\end{tabular}}  
\end{table}
}

\section{Conclusion}
We introduce \bench{}, the first benchmark dedicated to dependency inference across 581 repositories in four programming languages: Python, C\#, Rust, and JavaScript. In addition to measuring textual accuracy, we propose a novel CI-based evaluation that incorporates actual tests execution. Extensive experiments on various open-source and proprietary LLMs demonstrate that even the most advanced models struggle to infer dependencies accurately, highlighting opportunities for future advancements. We believe this study lays the groundwork for repository-level code development, with dependency inference serving as a pivotal step toward fully automated code generation.

\section*{Limitations}

Our study acknowledges several limitations. \ding{182}~Due to constraints in computing resources, our evaluation primarily focused on five mainstream models, selecting smaller model sizes. While these models are sufficiently representative, broadening the scope to include a greater variety of LLMs with diverse sizes could potentially enrich our findings. \ding{183}~In our experiments, we employed the GPT-4o and GPT-4o mini models, which operate as black boxes. The outputs may vary due to potential model upgrades or fluctuations in resources. To mitigate this issue, we provide the dates of the model versions used as a reference and set the temperature to 0 to ensure more consistent outputs. \ding{184}~Test coverage for each repository may not be exhaustive, meaning some test cases might not encompass every possible code path. However, as the tests were developed by project contributors, the results are expected to reflect practical settings accurately.


%% file: sections/7_appendix.tex
\section{Distribution of \bench{} Dataset on Token Count and Dependency Amount}
\begin{figure}[h]
    \centering
    \includegraphics[width=\linewidth]{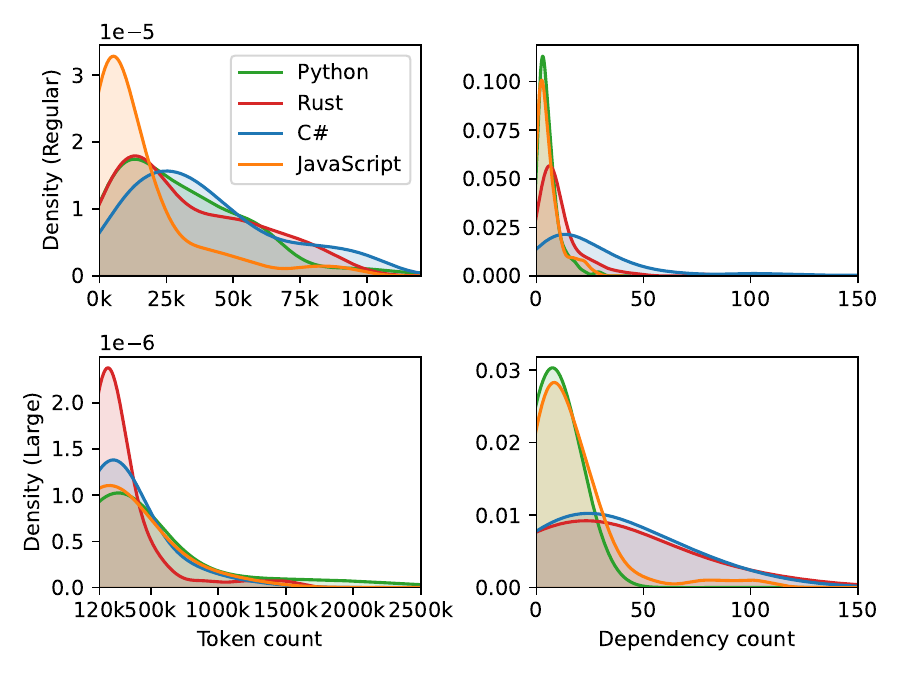}
    \caption{Distribution of token and dependency count.}
    \label{fig:distri}
\end{figure}

Figure~\ref{fig:distri} illustrates the distribution of token and dependency counts across different programming languages (Python, Rust, C\#, and JavaScript) for both Regular and Large repositories. For Regular repositories, the token count distribution shows that Python and Rust have a higher density at lower token counts, indicating that these languages typically have smaller codebases. In contrast, C\# and JavaScript display a more spread-out distribution, suggesting a wider range of codebase sizes. When examining Large repositories, the token count distribution shifts substantially, with all languages showing a lower density, highlighting the increased complexity and size of codebases in larger repositories.

The dependency count distribution for Regular repositories reveals that most dependencies are concentrated in the lower range across all languages, with Python and Rust having slightly higher densities at lower counts. For Large repositories, the dependency count distribution shows a similar pattern but with slightly higher densities for C\# and JavaScript, indicating these languages tend to have more dependencies in larger codebases.

\section{Additional Experiments}

\subsection{Performance of Baseline Methods with GPT-4o-mini}

\begin{table*}[h]  
\centering  
\caption{Performance of benchmark methods across programming languages and repository sizes on GPT-4o-mini (Continue to Table~\ref{tab:baseline-performance-repo-size} with a different model)}  
\label{tab:baseline-performance-repo-size-2}  
\resizebox{0.8\textwidth}{!}{  
\begin{tabular}{llcccccccccc}
\toprule  
\multirow{2}{*}{Lang} & \multirow{2}{*}{Method} & \multicolumn{5}{c}{Regular} & \multicolumn{5}{c}{Large} \\  
\cmidrule(lr){3-7} \cmidrule(lr){8-12}  
             &               & Exec & P & R & F1 & FR & Exec & P & R & F1 & FR \\  
\midrule  
\multirow{3}{*}{Python} & \baselineall{}     & 25.5 & 56.5 & 57.5 & 57.0 & 2.0 & - & - & - & - & -\\  
                        & \baselinefile{}     & 21.4 & 41.7 & 63.7 & 50.4 & 2.8 & 14.0 & 31.0 & 27.6 & 29.2 & 4.2 \\  
                        & \baselineimport{}   & 30.6 & 59.0 & 62.0 & 60.5 & 1.6 & 18.0 & 45.4 & 32.3 & 37.7 & 3.1 \\  
\midrule  
\multirow{3}{*}{Rust}   & \baselineall{}    & 7.1 & 76.0 & 49.1 & 59.7 & 1.0 & - & - & - & - & -\\  
         & \baselinefile{}    & 4.1 & 74.8 & 60.4 & 66.8 & 1.5 & 0.0 & 36.9 & 45.2 & 40.6 & 4.5   \\  
         & \baselineimport{}   & 1.0 & 77.7 & 49.4 & 60.4 & 1.0 & 0.0 & 64.1 & 23.8 & 34.7 & 4.0 \ \\  
\midrule    
\multirow{3}{*}{C\#}    & \baselineall{}    & 3.1 & 41.2 & 18.8 & 25.8 & 12.6 & - & - & - & - & - \\  
                         & \baselinefile{}   & 3.1 & 25.0 & 22.8 & 23.8 & 15.6  & 0.0 & 19.2 & 13.6 & 15.9 & 5.8 \\  
                         & \baselineimport{} & 3.1 & 44.2 & 23.5 & 30.7 & 6.7   & 0.0 & 34.7 & 14.1 & 20.1 & 6.2  \\  
\midrule  
\multirow{3}{*}{JavaScript}  
         & \baselineall{}    & 17.9 & 84.6 & 31.6 & 46.0 & 2.5  & - & - & - & - & - \\  
         & \baselinefile{}    & 16.8 & 45.7 & 25.8 & 33.0 & 7.2 & 2.2 & 27.0 & 20.2 & 23.1 & 3.5 \\   
         & \baselineimport{}   & 13.7 & 67.5 & 19.0 & 29.6 & 1.3  & 2.2 & 57.9 & 9.1 & 15.7 & 0.0  \\  
\bottomrule  
    \end{tabular}}  
\end{table*} 

\label{appendix:baseline-result}
Table~\ref{tab:baseline-performance-repo-size-2} presents the performance of various benchmark methods across different languages and repository sizes (Regular and Large) on GPT-4o-mini. 
Notably, the effectiveness of these methods varies significantly between Regular and Large repositories, with performance generally declining as repository size increases. Python and Rust show relatively higher performance in Regular repositories compared to C\# and JavaScript, which struggle more consistently across both repository sizes. Furthermore, the \baselineimport{} method for Python and \baselinefile{} method for Rust stand out with comparatively better performance in Regular repositories. The results indicate that while some methods perform well in smaller repositories, there is a significant drop in effectiveness in larger repositories, underscoring the importance of optimizing methods to handle different repository scales efficiently. The conclusion aligns with the findings we obtained in Section~\ref{sec:baseline_results}. Besides, the variability suggests that a one-size-fits-all approach is insufficient, and tailored strategies are necessary to maintain high performance across different contexts.

\subsection{Performance When Varying the Model Size}
\label{appendix:qwen-various-size}
\begin{figure}[!h]
    \centering
    \includegraphics[width=\linewidth]{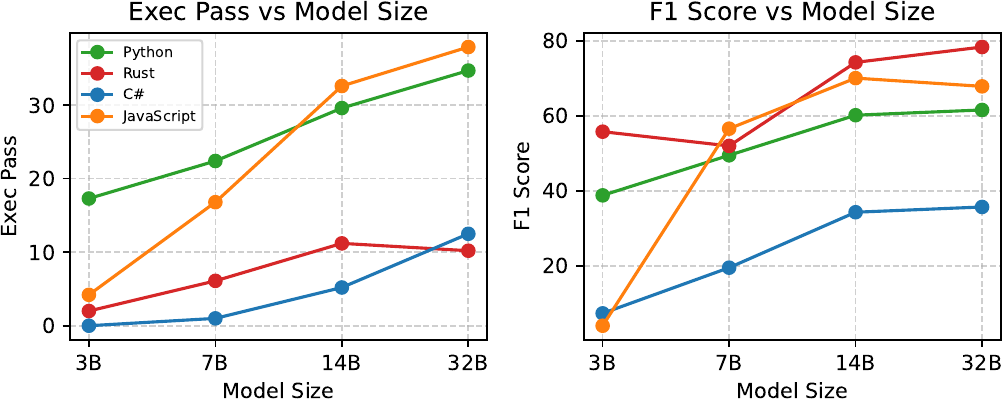}
    \caption{Model performance across programming languages for Qwen2.5-Coder-Instruct}
    \label{fig:perf-size}
\end{figure}

Figure~\ref{fig:perf-size} presents the performance of \baselineall{} approach on Regular dataset with different sizes of Qwen2.5-Coder-Instruct models. We observed a general trend where larger models consistently improved executability and textual accuracy metrics across four languages. Besides, when increasing the model size for compiled languages ike Rust and C\#, textual accuracy increases sharply, but the executability remains relative low, demonstrating the great value of our execution-based evaluation in benchmarking.

\subsection{Performance When Varying the Repository Size}
\label{appendix:repo_size}

\begin{figure}[!ht]
    \includegraphics[width=1\linewidth]{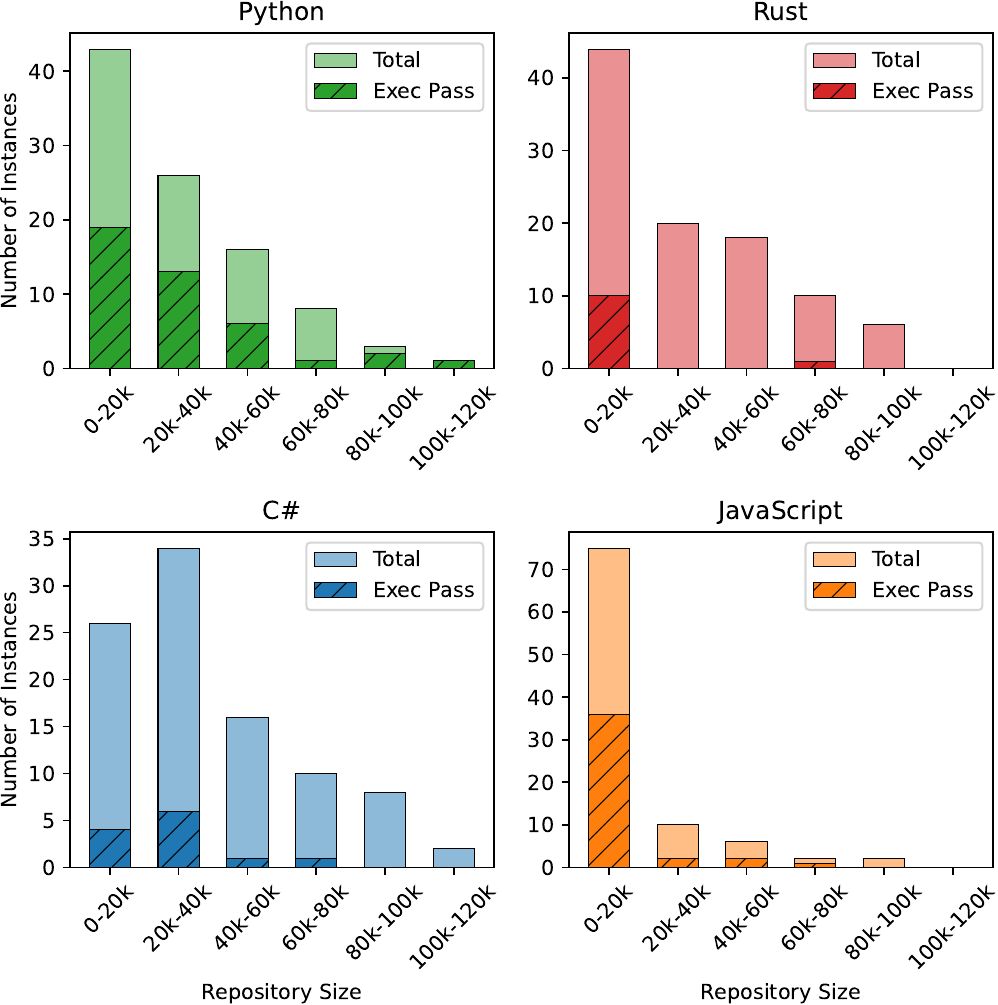}
    \caption{Execution pass rate w.r.t repository size}
    \label{fig:exec_token}
\end{figure}

In Table~\ref{tab:baseline-performance-repo-size}, we can observe that that dependency inference performance deteriorates when applied to larger datasets. However, whether this finding is generally applicable is unconfirmed. Therefore, we conducted a further analysis within the regular dataset which contains repositories of varying sizes. The results are depicted in Figure~\ref{fig:exec_token}, showing a decline in executability rates as repository size increases. Hence, there exists a negative correlation between repository size and model performance both between and within datasets. This suggests that long-context reasoning remains a significant challenge for LLMs, as longer input contexts lead to increased complexity in managing the project dependencies. This finding aligns with previous studies on long-context reasoning~\cite{ruler, longbench}.

\section{Example of Configuration Files}
\label{appdix:config}

This section introduces the types of configuration files for the four languages involved in this paper. These files specify project dependencies and serve as carriers for storing inference results. They also exercise the capabilities of LLMs to interact with modern programming languages build systems, it is important to provide a clear demonstration here.

\begin{figure}[h!]
    \centering
    \includegraphics[width=\linewidth]{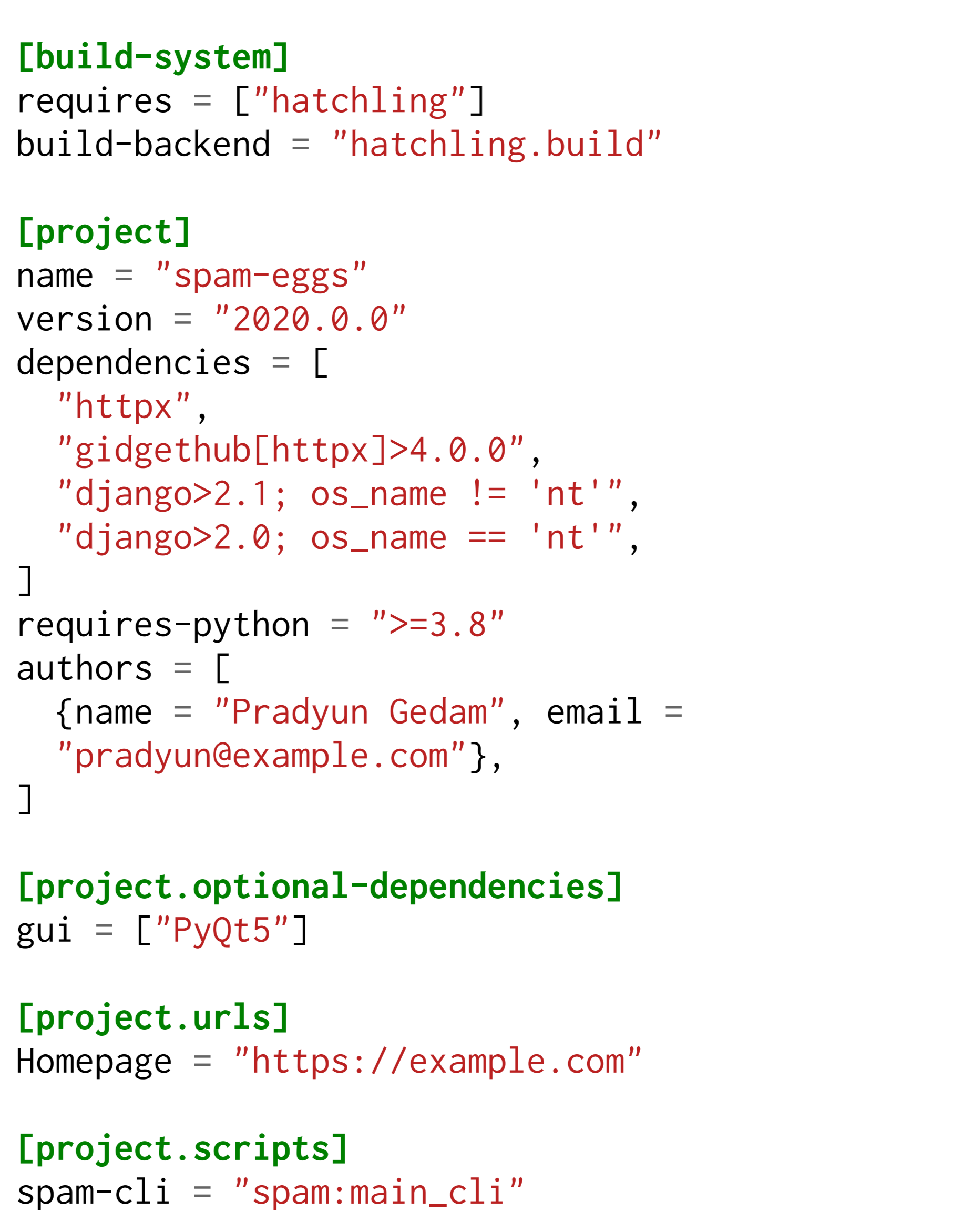}
    \caption{An example of \texttt{pyproject.toml} in Python}
    \label{fig:pyproject}
\end{figure}

\para{Python} (Figure~\ref{fig:pyproject})
\texttt{pyproject.toml} is the configuration file used by most Python projects. It includes sections for specifying metadata such as package names and authors, defining project dependencies, and configuring various development tools.

\begin{figure}[h!]
    \centering
    \includegraphics[width=\linewidth]{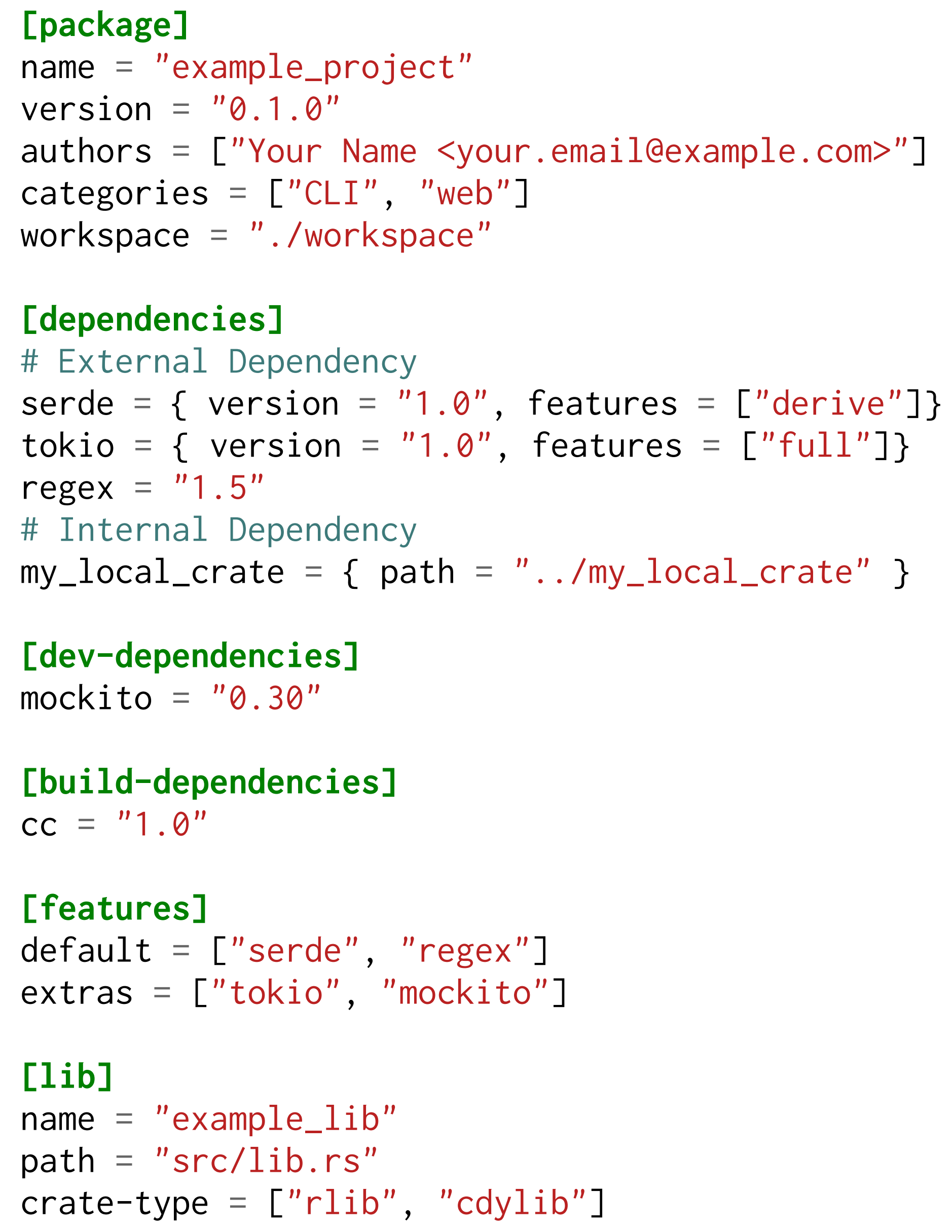}
    \caption{An example of \texttt{Cargo.toml} in Rust}
    \label{fig:cargo}
\end{figure}

\para{Rust} (Figure~\ref{fig:cargo})
\texttt{Cargo.toml} is the configuration file used in Rust projects. A single repository may contain multiple local crates, each with its own \texttt{Cargo.toml}, requiring proper configuration of internal dependency references.

\begin{figure}[h]
    \centering
    \includegraphics[width=\linewidth]{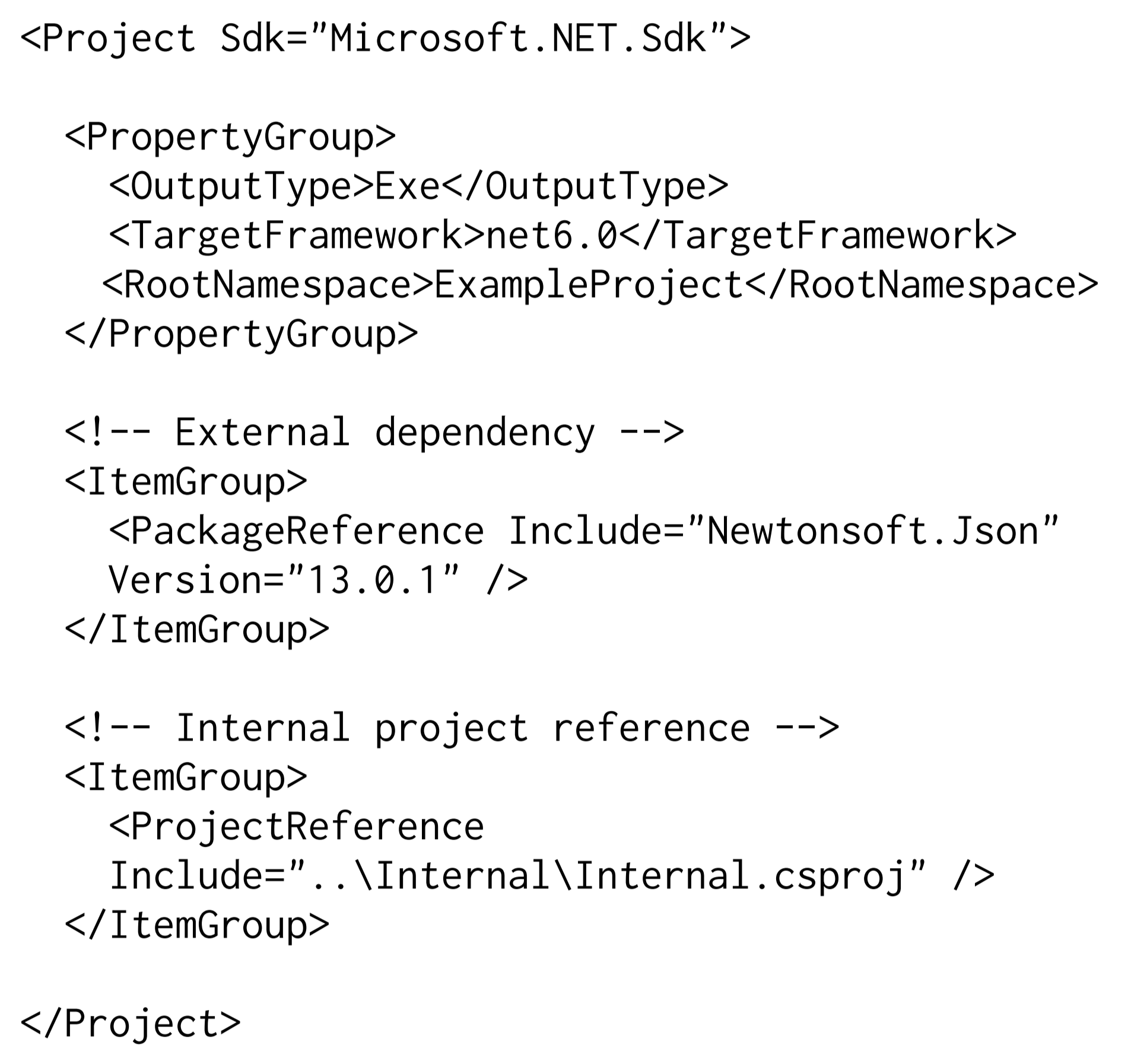}
    \caption{An example of \texttt{example.csproj} in C\#}
    \label{fig:csproj}
\end{figure}

\para{C\#} (Figure~\ref{fig:csproj})
Similar to Rust projects, C\# repositories are often structured as solutions containing multiple internal projects. Each project uses a \texttt{.csproj} configuration file to specify external and internal dependencies and configure compilation options.

\begin{figure}[!h]
    \centering
    \includegraphics[width=\linewidth]{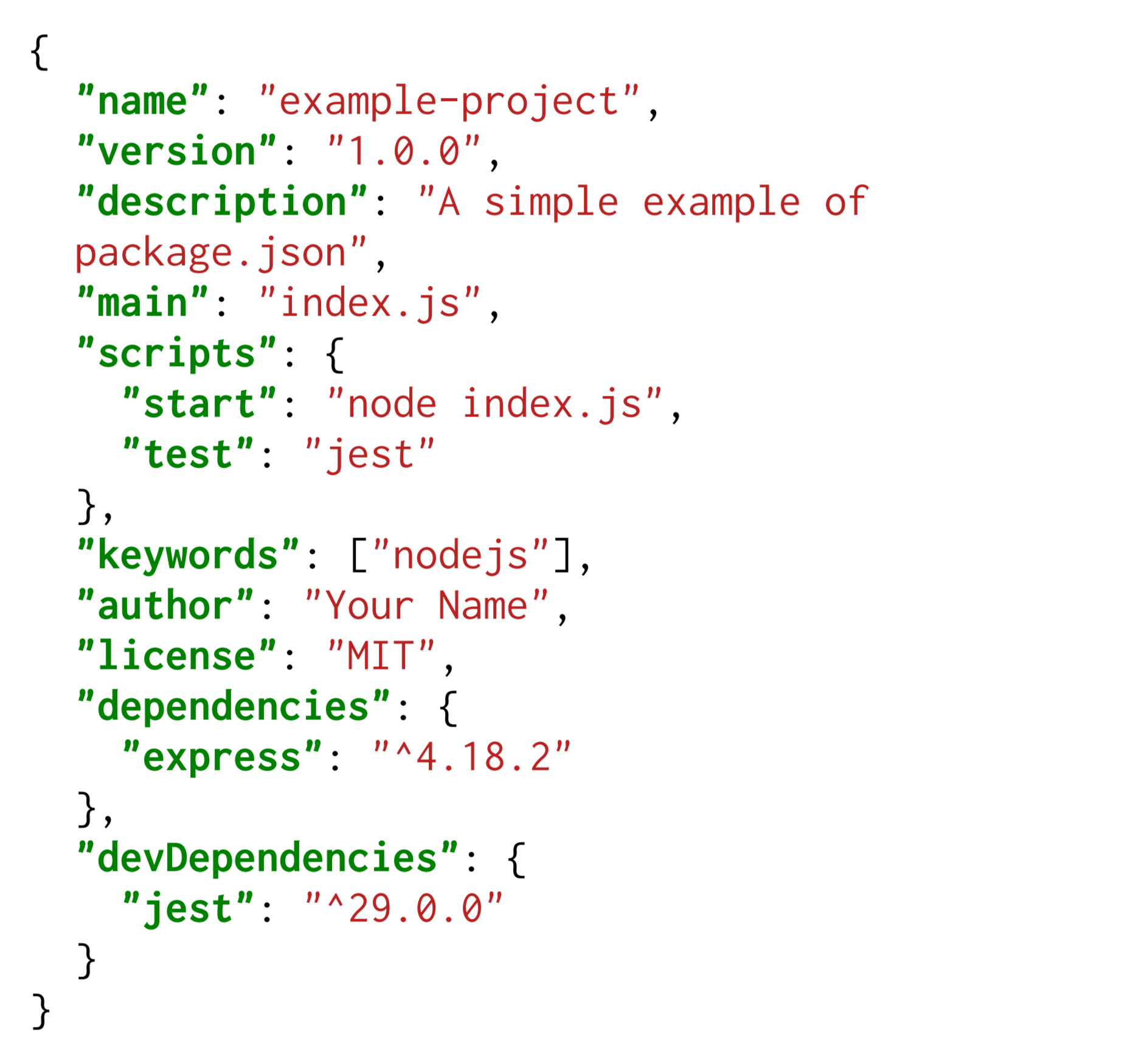}
    \caption{An example of \texttt{package.json} in JavaScript}
    \label{fig:packagejson}
\end{figure}

\para{JavaScript} (Figure~\ref{fig:packagejson})
\texttt{package.json} is the configuration file used in JavaScript projects, particularly those managed with Node.js. It defines metadata such as the project name, version, and description, and specifies dependencies, scripts, and entry points for the project. 

\vspace{.03in}
It can be found that dependency management constitutes the majority of the configuration files.

\section{Experimental Settings}

\subsection{Baseline \baselineall{}}
\label{appendix:all-in-one}
In \baselineall{}, our baseline approach feeds the entire codebase as input context to the LLM and processes the task through a single LLM call. The model simultaneously generates all build configurations, which are then parsed to obtain the updated build files. The complete prompt template used for this approach is detailed in Figure~\ref{fig:all_in_one_prompt}.

\label{appendix:prompt}

\begin{figure*}[h!]  
    \centering 
    \noindent
\begin{lstlisting}[language={},breakautoindent=false,breakindent=0ex]
Edit the build files to include all necessary dependency-related configurations to ensure the project builds and runs successfully. Output a copy of each build file.

You will receive four sections of information to configure dependencies in build files:
1. **Project Structure**: A tree structure representing the project's layout.
2. **Environment Specifications**: Details about the operating system and language SDK where the project will run.
3. **Source Code**: The full source code of the project.
4. **Build Files**: Build files missing dependency configurations, which you will need to update.

!Important Notes:
1. The project may include multiple build files. Ensure you update all of them with the necessary dependency configurations.
2. Only edit the files listed in the "Build Files" section.
3. Limit your edits strictly to dependency configurations within the build files.

To suggest changes to a file you MUST return the entire content of the updated file.
You MUST use this *file listing* format:

path/to/filename.js
```
// entire file content ...
// ... goes in between
```

Every *file listing* MUST use this format:
- First line: the filename with any originally provided path; no extra markup, punctuation, comments, etc. **JUST** the filename with path.
- Second line: opening ```
- ... entire content of the file ...
- Final line: closing ```

To suggest changes to a file you MUST return a *file listing* that contains the entire content of the file.
*NEVER* skip, omit or elide content from a *file listing* using "..." or by adding comments like "... rest of code..."!
Create a new file you MUST return a *file listing* which includes an appropriate filename, including any appropriate path.

--- Begin of Project Structure ---
{project_structure}
--- End of Project Structure ---

--- Begin of Environment Specifications ---
{env_specs}
--- End of Environment Specifications ---

--- Begin of Source Code ---
{src_section}
--- End of Source Code ---

--- Begin of Build Files ---
{build_section}
--- End of Build Files ---
\end{lstlisting} 
    \caption{Prompt template used to generate build file.}
    \label{fig:all_in_one_prompt}
\end{figure*}  

\subsection{Baseline \baselineimport{}}
\baselineimport{} follows the same prompting strategy as \baselineall{} with a single LLM call. The key distinction lies in the input composition: while \baselineall{} includes the complete codebase, \baselineimport{} only incorporates the import statements from source files in the input context. This selective approach focuses the model's attention on the most dependency-relevant code segments. We leverage tree-sitter to extract import statements across different programming languages, with detailed usage information provided in Appendix~\ref{appdix:treesitter}.

\subsection{Baseline \baselinefile{}}
\baselinefile{} employs a two-stage prompting strategy. In the first stage, it processes source files individually, applying the same prompt template which is detailed in Appendix~\ref{appendix:prompt} as previous baselines but with a single file as context per LLM call. This generates separate build files edits for each source file. In the second stage, for each build file, we merge its various updates from the first stage using a dedicated LLM call, where the prompt is shown in Figure~\ref{fig:merge_prompt}. The merge prompt template is detailed in Appendix~\ref{appdix:prompt_merge}. The final output consists of the comprehensively updated build files derived from this two-stage process.

\label{appdix:prompt_merge}

\begin{figure*}[h!]  
    \centering 
    \noindent
\begin{lstlisting}[language={},breakautoindent=false,breakindent=0ex]
Here is a list of edits to a project's build files, which is generated by add dependency configuration according to each source file. Edit the build files to merge all edits in the "Build File Edits" section to ensure the project builds and runs successfully. Output a copy of the build file.

You will receive four sections of information to configure dependencies in build files:
1. **Project Structure**: A tree structure representing the project's layout.
2. **Environment Specifications**: Details about the operating system and language SDK where the project will run.
3. **Build File Edits**: A list of edited build file, which you will need to merge.
4. **Build File**: Build files missing dependency configurations, which you will need to update based on above edits.

To suggest changes to a file you MUST return the entire content of the updated file.
You MUST use this *file listing* format:

path/to/filename.js
```
// entire file content ...
// ... goes in between
```

Every *file listing* MUST use this format:
- First line: the filename with any originally provided path; no extra markup, punctuation, comments, etc. **JUST** the filename with path.
- Second line: opening ```
- ... entire content of the file ...
- Final line: closing ```

To suggest changes to a file you MUST return a *file listing* that contains the entire content of the file.
*NEVER* skip, omit or elide content from a *file listing* using "..." or by adding comments like "... rest of code..."!
Create a new file you MUST return a *file listing* which includes an appropriate filename, including any appropriate path.

--- Begin of Project Structure ---
{project_structure}
--- End of Project Structure ---

--- Begin of Environment Specifications ---
{env_specs}
--- End of Environment Specifications ---

--- Begin of Build File Edits---
{build_file_edits}
--- End of Build Files Edits---

--- Begin of Build File ---
{build_section}
--- End of Build File ---
\end{lstlisting} 
    \caption{Prompt used to merge build file edits.}
    \label{fig:merge_prompt}
\end{figure*}  

\subsection{Tree-sitter}
\label{appdix:treesitter}

Tree-sitter is a parsing system widely used in code analysis that generates concrete syntax trees for source code.  In our implementation, we utilize Tree-sitter to extract import statements and dependency-related code segments across different programming languages. Tree-sitter's language-agnostic nature and robust parsing capabilities enable our system to maintain consistent analysis quality across Python, JavaScript, Rust, and other supported languages. Tree-sitter queries provide a powerful pattern-matching language for searching syntax trees. The query language allows precise targeting of syntax tree patterns using a declarative, S-expression-based syntax. Below is the queries we used to extract import statements. 

\begin{lstlisting}[language={python},breakautoindent=false,breakindent=0ex]
# Python
[(import_statement) (import_from_statement)] @import

# Rust
(use_declaration) @use

# C#
(using_directive) @use

# JavaScript
(import_statement) @import
\end{lstlisting} 

\subsection{Model Serving}
For GPT-4o and GPT-4o-mini, we utilize the specific versions gpt-4o-20240806 and gpt-4o-mini-20240718, accessed through the OpenAI API.  For open-source models, we employ checkpoints available on Hugging Face and serve them using VLLM across 4 A100 GPUs. The decoding strategy is configured as greedy decoding with a maximum output token limit of 8,000.